\documentclass[final]{cvpr}

\usepackage[utf8]{inputenc} % allow utf-8 input
\usepackage[T1]{fontenc}    % use 8-bit T1 fonts
\usepackage{url}            % simple URL typesetting
\usepackage{booktabs}       % professional-quality tables
\usepackage{amsfonts}       % blackboard math symbols
\usepackage{nicefrac}       % compact symbols for 1/2, etc.
\usepackage{microtype}      % microtypography
\usepackage{color}
\usepackage{times}
\usepackage{epsfig}
\usepackage{graphicx}
\usepackage{amsmath}
\usepackage{amssymb}
\usepackage{algorithm}
\usepackage{multirow}
\usepackage{makecell}
\usepackage{tabularx}
\usepackage{bbm}
\usepackage{algorithmic}
\usepackage{enumitem}
\usepackage{dsfont}
\newcolumntype{L}[1]{>{\raggedright\arraybackslash}p{#1}}
\newcolumntype{C}[1]{>{\centering\arraybackslash}p{#1}}
\newcolumntype{R}[1]{>{\raggedleft\arraybackslash}p{#1}}

\newcommand{\z}{{\mathbf{z}}}
\newcommand{\x}{{\mathbf{x}}}

\newcommand*\samethanks[1][\value{footnote}]{\footnotemark[#1]}

\setlist[itemize]{leftmargin=4.mm}

\usepackage[pagebackref=true,breaklinks=true,colorlinks,bookmarks=false]{hyperref}

\begin{document}

%%%%%%%%% TITLE
\title{Line Segment Detection Using Transformers without Edges}

\author{Yifan Xu\thanks{~indicates equal contribution.}, ~~Weijian Xu\samethanks, ~~David Cheung, ~~Zhuowen Tu \\
University of California San Diego \\
% Institution1 address\\
{\tt\small $\{$yix081,wex041,d6cheung,ztu$\}$@ucsd.edu }
}

\maketitle

%%%%%%%%% ABSTRACT
\begin{abstract}
In this paper, we present a joint end-to-end line segment detection algorithm using Transformers that is post-processing and heuristics-guided intermediate processing (edge/junction/region detection) free. Our method, named LinE segment TRansformers (LETR), takes advantages of having integrated tokenized queries, a self-attention mechanism, and encoding-decoding strategy within Transformers by skipping standard heuristic designs for the edge element detection and perceptual grouping processes. We equip Transformers with a multi-scale encoder/decoder strategy to perform fine-grained line segment detection under a direct endpoint distance loss. This loss term is particularly suitable for detecting geometric structures such as line segments that are not conveniently represented by the standard bounding box representations. The Transformers learn to gradually refine line segments through layers of self-attention. In our experiments, we show state-of-the-art results on Wireframe and YorkUrban benchmarks. 
\let\thefootnote\relax\footnotetext{Code: \url{https://github.com/mlpc-ucsd/LETR}.}

\end{abstract}

%%%%%%%%% BODY TEXT
\section{Introduction}

Line segment detection is an important mid-level visual process \cite{marr1982vision} useful for solving various downstream computer vision tasks, including segmentation, 3D reconstruction, image matching and registration, depth estimation, scene understanding, object detection, image editing, and shape analysis. Despite its practical and scientific importance, line segment detection remains an unsolved problem in computer vision.

\begin{figure}[!htp]
\centering
\label{fig:pipeline-comparison}
\begin{tabular} {c}
\hspace{-1mm}\includegraphics[width=0.47\textwidth]{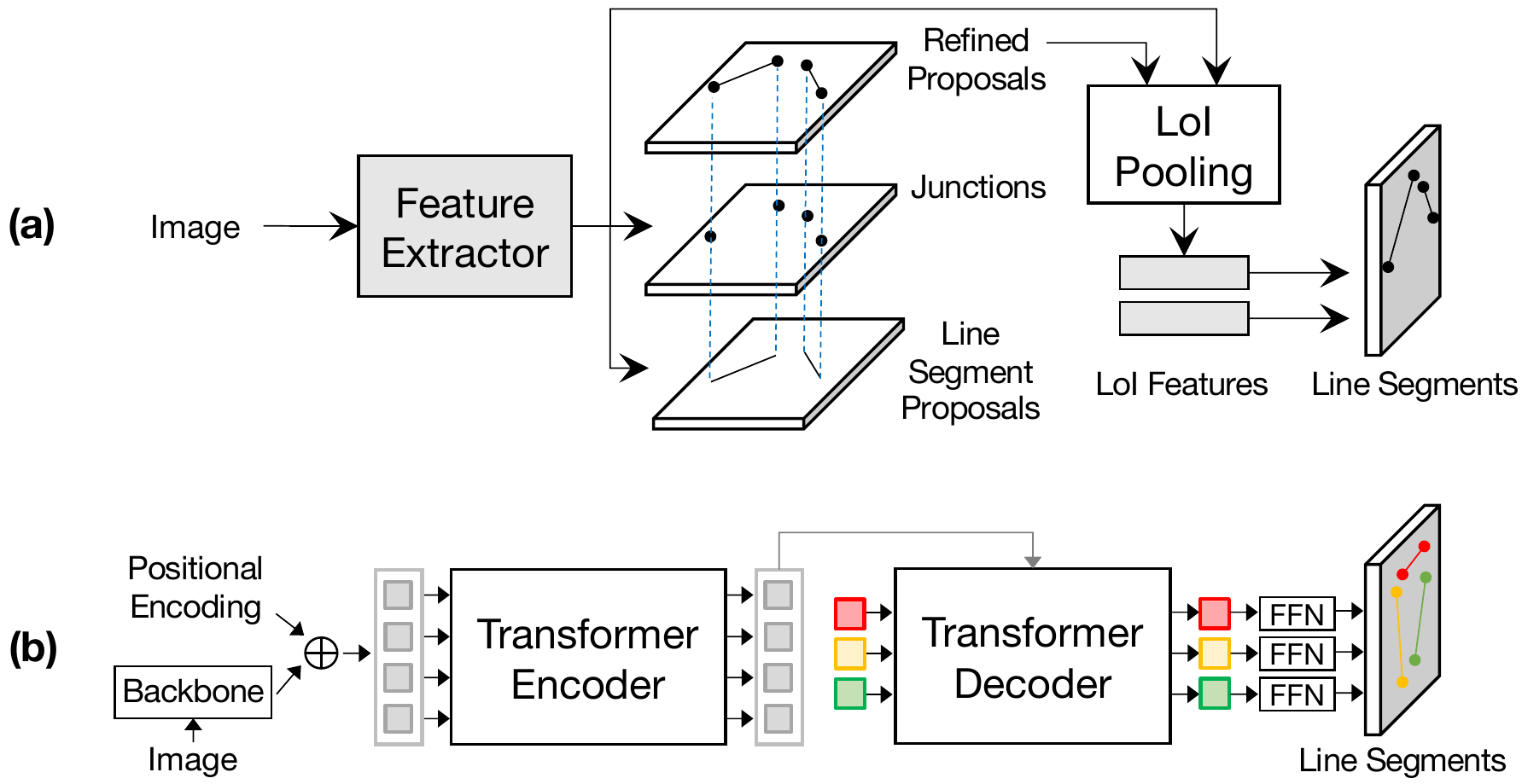}
\end{tabular}
\caption{\footnotesize \textbf{Pipeline comparison} between: (a) holistically-attracted wireframe parsing (HAWP) \cite{xue2020holistically} and (b) our proposed LinE segment TRansformers (LETR). LETR is based on a general-purpose pipeline without heuristics-driven intermediate stages for detecting junctions and generating line segment proposals.}
\vspace{-5mm}
\end{figure}

Although dense pixel-wise edge detection has achieved an impressive performance \cite{xie2015holistically}, reliably extracting line segments of semantic and perceptual significance remains a further challenge. In natural scenes, line segments of interest often have heterogeneous structures within the cluttered background that are locally ambiguous or partially occluded. Morphological operators \cite{smith1997susan} operated on detected edges \cite{canny1986computational} often give sub-optimal results. Mid-level representations such as Gestalt laws \cite{elder2002ecological} and contextual information \cite{tu2008auto} can play an important role in the perceptual grouping, but they are often hard to be seamlessly integrated into an end-to-end line segment detection pipeline. Deep learning techniques \cite{krizhevsky2012imagenet,long2015fully,he2016deep,xie2015holistically} have provided greatly enhanced feature representation power, and algorithms such as \cite{zhou2019end,xue2019learning,xue2020holistically} become increasingly feasible in real-world applications.
However, systems like \cite{zhou2019end,xue2019learning,xue2020holistically} still consist of heuristics-guided modules \cite{smith1997susan} such as edge/junction/region detection, line grouping, and post-processing, limiting the scope of their performance enhancement and further development.

In this paper, we skip the traditional edge/junction/region detection + proposals + perceptual grouping pipeline by designing a Transformer-based \cite{vaswani2017attention,carion2020end} joint end-to-end line segment detection algorithm. We are motivated by the following observations for the Transformer frameworks \cite{vaswani2017attention,carion2020end}: tokenized queries with an integrated encoding and decoding strategy, self-attention mechanism, and bipartite (Hungarian) matching step, capable of addressing the challenges in line segment detection for edge element detection, perceptual grouping, and set prediction; general-purpose pipelines for Transformers that are heuristics free. Our system, named LinE segment TRsformer (LETR), enjoys the modeling power of a general-purpose Transformer architecture while having its own enhanced property for detecting fine-grained geometric structures like line segments. LETR is built on top of a seminal work, DEtection TRansformer (DETR) \cite{carion2020end}. However, as shown in Section \ref{results_comparison} for ablation studies, directly applying the DETR object detector \cite{carion2020end} for line segment detection does not yield satisfactory results since line segments are elongated geometric structures that are not feasible for the bounding box representations.

Our contributions are summarized as follows.
\begin{itemize}
 \setlength\itemsep{0mm}
 \setlength{\itemindent}{0mm}
    \item We cast the line segment detection problem in a joint end-to-end fashion without explicit edge/junction/region detection and heuristics-guided perceptual grouping processes, which is in distinction to the existing literature in this domain. We achieve state-of-the-art results on the Wireframe \cite{huang2018learning} and YorkUrban benchmarks \cite{denis2008efficient}. 
    \item We perform line segment detection using Transformers, based specifically on DETR \cite{carion2020end}, to realize tokenized entity modeling, perceptual grouping, and joint detection via an integrated encoder-decoder, a self-attention mechanism, and joint query inference within Transformers.
    \item We introduce two new algorithmic aspects to DETR \cite{carion2020end}: first, a multi-scale encoder/decoder strategy as shown in Figure \ref{fig:pipeline}; second, a direct endpoint distance loss term in training, allowing geometric structures like line segments to be directly learned and detected --- something not feasible in the standard DETR bounding box representations.
\end{itemize}

\section{Related Works}

\subsection{Line Segment Detection}
\vspace{-2mm}
\paragraph{Traditional Approaches.} Line detection has a long history in computer vision. Early pioneering works rely on low-level cues from pre-defined features (e.g. image gradients). 
Typically, line (segment) detection performs edge detection \cite{canny1986computational,martin2004learning,dollar2006supervised,dollar2013structured,xie2015holistically}, followed by a perceptual grouping \cite{guil1995fast,smith1997susan,elder2002ecological} process.
Classic \textit{perceptual grouping} frameworks \cite{BurnsHR86,boldt1989token,nieto2011line,lu2015cannylines,VonGioi2010} aggregate the low-level cues to form line segments in a bottom-up fashion: an image is partitioned into line-support regions by grouping similar pixel-wise features. Line segments are then approximated from line-support regions and filtered by a validation step to remove false positives. Another popular series of line segment detection approaches are based on \textit{Hough transform} \cite{duda1972use,guil1995fast,matas2000robust,furukawa2003accurate} by gathering votes in the parameter space: the pixel-wise edge map of an image is converted into a parameter space representation, in which each point corresponds to a unique parameterized line. The points in the parameter space that accumulate sufficient votes from the candidate edge pixels are identified as line predictions. However, due to the limitations in the modeling/inference processes, these traditional approaches often produce sub-optimal results.
\vspace{-4mm}
\paragraph{Deep Learning Based Approaches.} The recent surge of deep learning based approaches has achieved much-improved performance on the line segment detection problem \cite{huang2018learning,xue2019learning,zhou2019end,PPGNet,xue2020holistically} with the use of learnable features to capture extensive context information.
% Junction-based approaches.
One typical family of methods is \textit{junction-based pipelines}: Deep Wireframe Parser (DWP) \cite{huang2018learning} creates two parallel branches to predict the junction heatmap and the line heatmap, followed by a merging procedure. Motivated by \cite{fasterrcnn}, L-CNN \cite{zhou2019end} simplifies \cite{huang2018learning} into a unified network. First, a junction proposal module produces the junction heatmap and then converts detected junctions into line proposals. Second, a line verification module classifies proposals and removes unwanted false-positive lines. Methods like \cite{zhou2019end} are end-to-end, but they are at the instance-level (for detecting the individual line segments). Our LETR, like DETR \cite{carion2020end}, has a general-purpose architecture that is trained in a holistically end-to-end fashion. PPGNet \cite{PPGNet} proposes to create a point-set graph with junctions as vertices and model line segments as edges. However, the aforementioned approaches are heavily dependent on high-quality junction detection, which is error-prone to various imaging conditions and complex scenarios. 

% Dense prediction based approaches.
Another line of approaches employs \textit{dense prediction} to obtain a surrogate representation map and applies a post-process procedure to extract line segments: AFM \cite{xue2019learning} proposes an attraction field map as an intermediate representation that contains 2-D projection vectors pointing to associated lines. A squeeze module then recovers vectorized line segments from the attraction field map. Despite a relatively simpler design, \cite{xue2019learning} demonstrates its inferior performance compared with junction-based approaches. Recently, HAWP \cite{xue2020holistically} builds a hybrid model of AFM \cite{xue2019learning}, and L-CNN \cite{zhou2019end} by computing line segment proposals from the attraction field map and then refining proposals with junctions before further line verification. 

In contrast, as shown in Figure \ref{fig:pipeline-comparison}, our approach differs from previous methods by removing heuristics-driven intermediate stages for detecting edge/junction/region proposals and surrogate prediction maps. Our approach is able to directly predict vectorized line segments while keeping competitive performances under a general-purpose framework.

\begin{figure*}
\centering
\includegraphics[width=18cm]{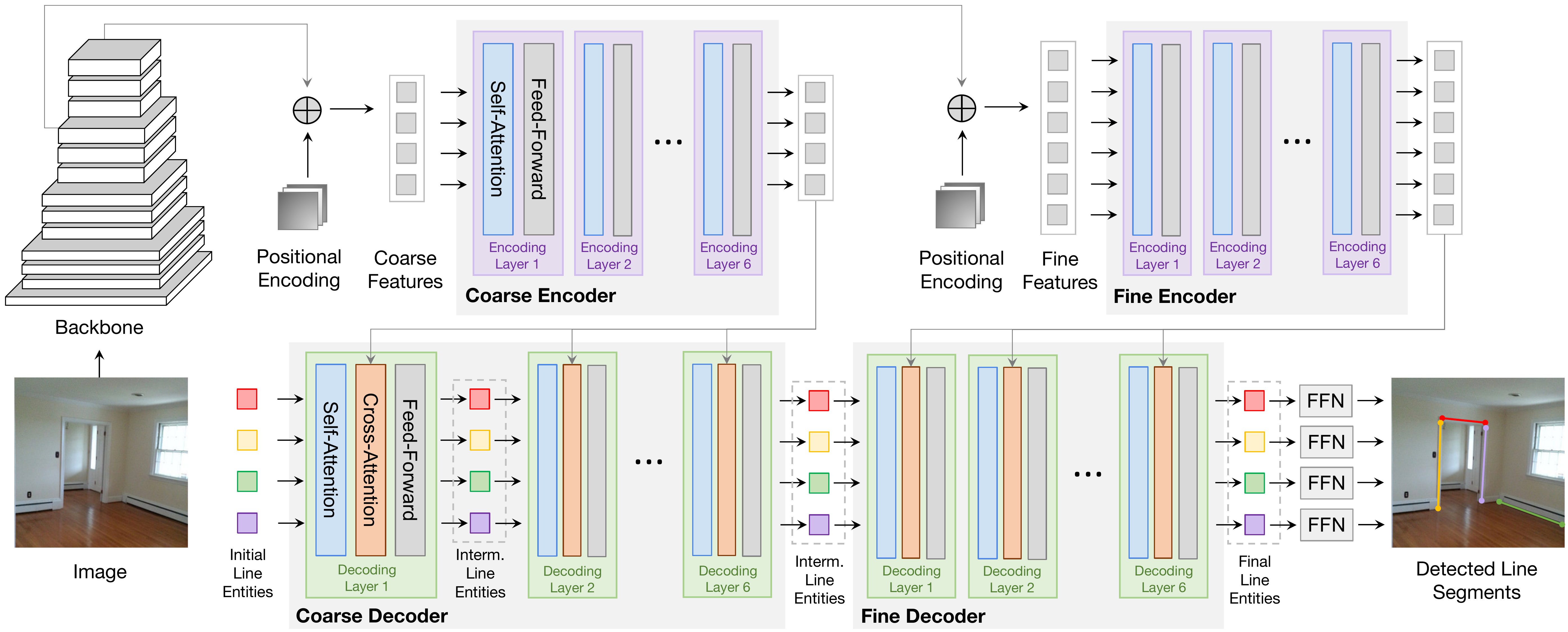}
\vspace{-0.2cm}
\caption{\textbf{Schematic illustration of our LETR pipeline}: An image is fed into a backbone network and generates two feature maps, which are then used by the coarse and the fine encoder respectively. Initial line entities are then first refined by the coarse decoder with the interaction of the coarse encoder output, and then the intermediate line entities from the coarse decoder are further refined by the fine decoder attending to the fine encoder. Finally, line segments are detected by feed-forward networks (FFNs) on top of line entities.}
\label{model}
\label{fig:pipeline}
\end{figure*}

\subsection{Transformer Architecture}
\vspace{-2mm}
Transformers \cite{vaswani2017attention} have achieved great success in the natural language processing field and become \textit{de facto} standard backbone architecture for many language models \cite{vaswani2017attention,devlin2018bert}. It introduces self-attention and cross-attention modules as basic building blocks, modeling dense relations among elements of the input sequence. These attention-based mechanisms also benefit many vision tasks such as video classification \cite{wang2018non}, semantic segmentation \cite{fu2019dual}, image generation \cite{zhang2019self}, etc. Recently, end-to-end object detection with Transformers (DETR) \cite{carion2020end} reformulates the object detection pipeline with Transformers by eliminating the need for hand-crafted anchor boxes and non-maximum suppression steps. Instead, \cite{carion2020end} proposes to feed a set of object queries into the encoder-decoder architecture with interactions from the image feature sequence and generate a final set of predictions. A bipartite matching objective is then optimized to force unique assignments between predictions and targets. 

We introduce two new aspects to DETR \cite{carion2020end} when realizing our LETR: 1) multi-scale encoder and decoder; 2) direct distance loss for the line segments.

\section{Line Segment Detection with Transformers}

\subsection{Motivation}
\vspace{-2mm}
\label{motivation}
\begin{figure}[!htp]
\vspace{-1mm}
\begin{center}
\begin{tabular} {c}
\hspace{-1mm}\includegraphics[width=0.40\textwidth]{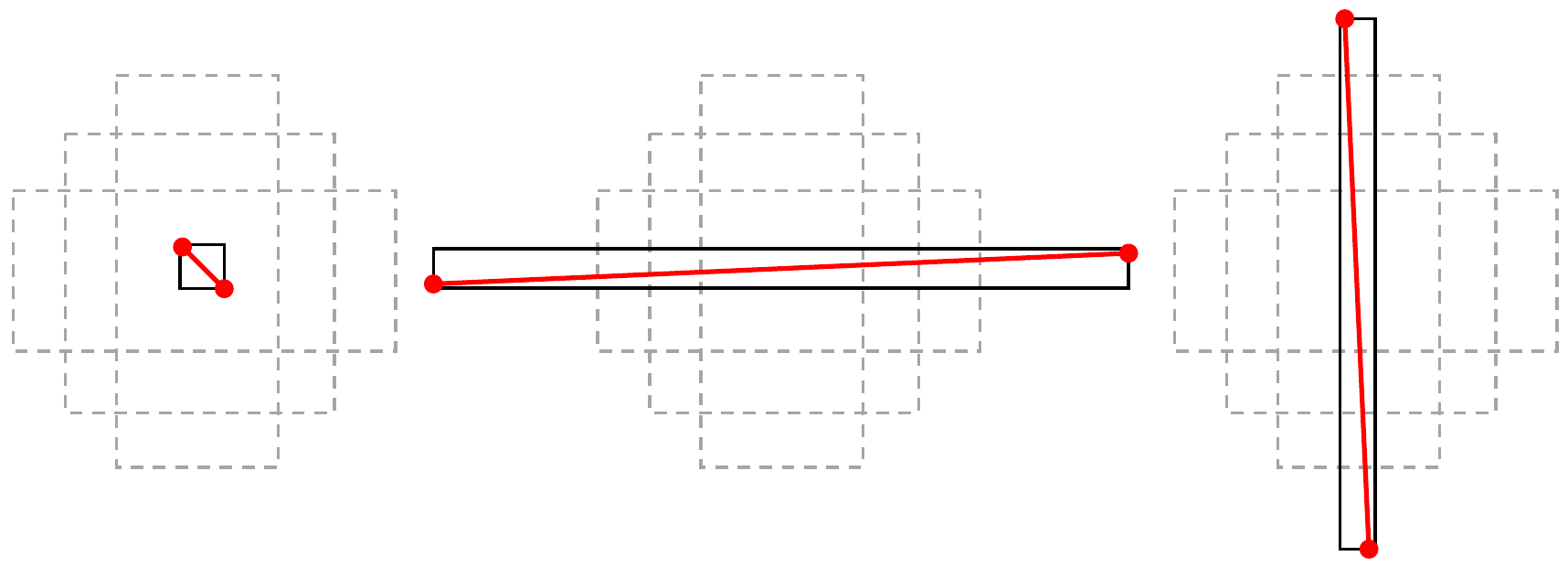}
\end{tabular}
\end{center}
\vspace{-1mm}
\caption{\footnotesize \textbf{Bounding box representation.} Three difficult cases to represent line segments using bounding box diagonals. Red lines, black boxes, and gray dotted boxes refer to as line segments, the corresponding bounding boxes, and anchors respectively.}
\label{fig:hard-cases-for-fasterrcnn}
\vspace{-4mm}
\end{figure}

Despite the exceptional performance achieved by the recent deep learning based approaches \cite{zhou2019end,xue2019learning,xue2020holistically} on line segment detection, their pipelines still involve heuristics-driven intermediate representations such as junctions and attraction field maps, raising an interesting question: \textit{Can we directly model all the vectorized line segments with a neural network?} A naive solution could be simply regarding the line segments as \textit{objects} and building a pipeline following the standard object detection approaches \cite{fasterrcnn}. Since the location of 2-D objects is typically parameterized as a bounding box, the vectorized line segment can be directly read from a diagonal of the bounding box associated with the line segment object. However, the limited choices of anchors make it difficult for standard two-stage object detectors to predict very short line segments or line segments nearly parallel to the axes (see Figure \ref{fig:hard-cases-for-fasterrcnn}). The recently appeared DETR \cite{carion2020end} eliminates the anchors and the non-maximum suppression, perfectly meets the need of line segment detection. However, the vanilla DETR still focuses on bounding box representation with a GIoU loss. We further convert the box predictor in DETR into a vectorized line segment predictor by adapting the losses and enhancing the use of multi-scale features in our designed model.

\begin{figure}
\begin{center}
\begin{tabular}{c}
\includegraphics[width=0.45\textwidth]{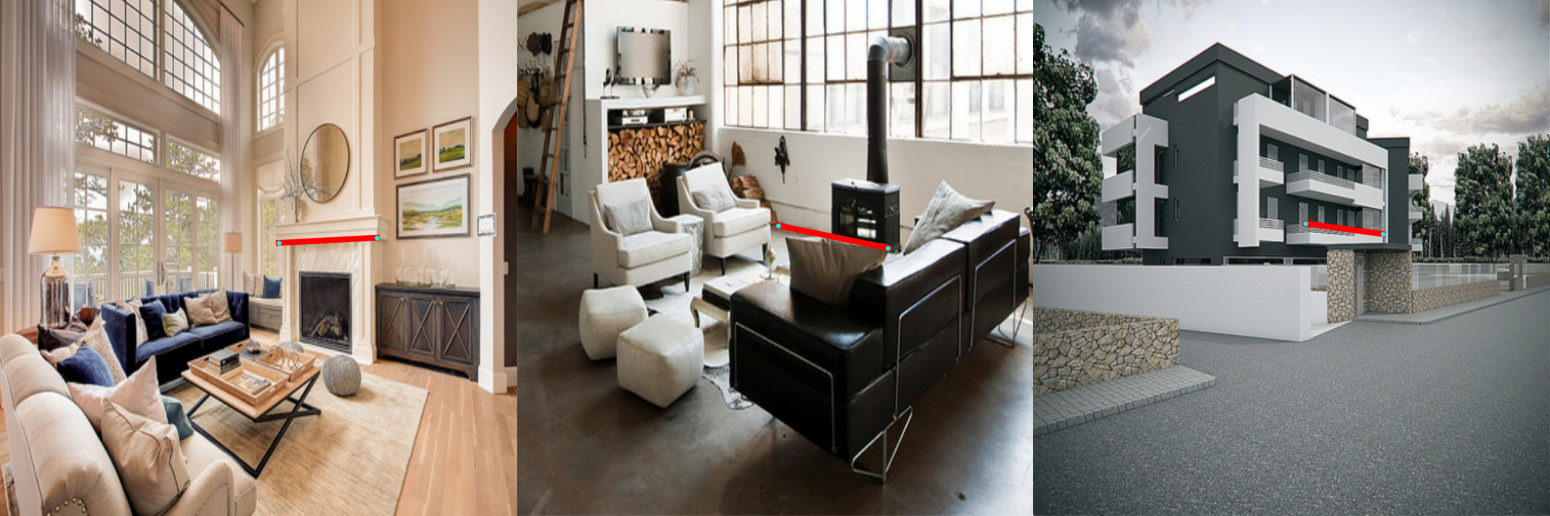} \\
\includegraphics[width=0.45\textwidth]{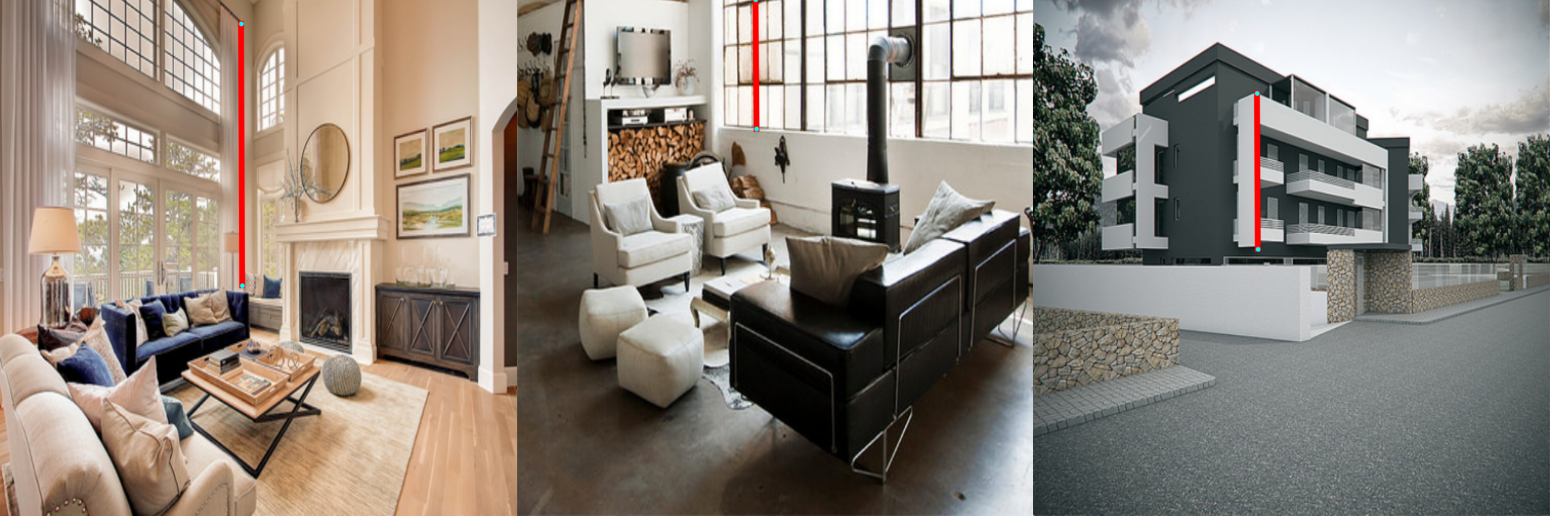}

\end{tabular}
\caption{\small \textbf{Line entity representation.} For each row, we show how a same line entity predicts line segments with same property in three different indoor/outdoor scenes. The top line entity is specialized for horizontal line segments in the middle of the figure, and the bottom one prefers to predict vertical line segments with a various range of lengths.}
\label{fig:vis_q1}
\vspace{-10mm}
\end{center}
\end{figure}
\vspace{-2mm}
\subsection{Overview}
\vspace{-2mm}
In a line segment detection task, a detector aims to predict a set of line segments from given images. Performing line segment detection with Transformers removes the need of explicit edge/junction/region detection \cite{zhou2019end, xue2020holistically} (see Figure \ref{fig:pipeline-comparison}). Our LETR is built purely based on the Transformer encoder-decoder structure. The proposed line segment detection process consists of four stages: \\
(1) {\em Image Feature Extraction}: Given an image input, we obtain the image feature map $\x \in \mathbb{R}^{H \times W \times C}$ from a CNN backbone with reduced dimension. The image feature is concatenated with positional embeddings to obtain spatial relations. (2) {\em Image Feature Encoding:} The flattened feature map $\x \in \mathbb{R}^{HW \times C}$ is then encoded to $\x' \in \mathbb{R}^{HW \times C}$ by a multi-head self-attention module and a feed forward network module following the standard Transformer encoding architecture. (3) {\em Line Segment Detection:} In the Transformer decoder networks, $N$ learnable line entities $\mathbf{l} \in \mathbb{R}^{N \times C}$ interact with the encoder output via the cross-attention module. (4) {\em Line Segment Prediction:} Line entities make line segment predictions with two prediction heads built on top of the Transformer decoder. The line coordinates are predicted by a multi-layer perceptron (MLP), and the prediction confidences are scored by a linear layer. 

\vspace{2mm}
\noindent\textbf{Self-Attention and Cross-Attention.} We first visit the scaled dot-product attention popularized by Transformer architectures \cite{vaswani2017attention}. The basic scaled dot-product attention consists of a set of $m$ queries $Q \in \mathbb{R}^{m \times d}$, a set of $n$ key-value pairs notated as a key matrix $K \in \mathbb{R}^{n \times d}$ and a value matrix $V \in \mathbb{R}^{n \times d}$. Here we set $Q$, $K$, $V$ to have same feature dimension $d$. 
The attention operation $F$ is defined as:
\vspace{-3mm}
\begin{equation}\label{eq:att}
\begin{split}
F&=\text{Att}(Q, K, V) = \text{softmax}(\frac{QK^{T}}{\sqrt{d}})V
\end{split}
\vspace{-3mm}
\end{equation}

In our encoder-decoder Transformer architecture, we adopt two attention modules based on the multi-head attention, namely the self-attention module (SA) and cross-attention (CA) module (see Figure \ref{fig:pipeline}). The SA module takes in a set of input embeddings notated as $\x = [x_{1}, ..., x_{i}] \in \mathbb{R}^{i \times d}$, and outputs a weighted summation $\x' = [x'_{1}, ..., x'_{i}] \in \mathbb{R}^{i \times d}$ of input embeddings within $\x$ following Eq.\ref{eq:att} where $F =\text{Att}(Q=\x,K=\x,V=\x)$. The CA module takes in two sets of input embeddings notated as $\x = [x_{1}, ..., x_{i}] \in \mathbb{R}^{i \times d}$, $\z = [x_{1}, ..., x_{j}] \in \mathbb{R}^{j \times d}$ following Eq.\ref{eq:att} where $F=\text{Att}(Q=\z,K=\x,V=\x)$.       

\noindent\textbf{Transformer Encoder in LETR} is stacked with multiple encoder layers. Each encoder layer takes in image features $\x \in \mathbb{R}^{HW \times c}$ from its predecessor encoder layer and processes it with a SA module to learn the pairwise relation. The output features from SA module are passed into a point-wise fully-connected layer (FC) with activation and dropout layer followed by another point-wise fully-connected (FC) layer. Layer norm is applied between SA module and first FC layer and after second FC layer. Residual connection is added before the first FC layer and after the second FC layer to facilitate optimization of deep layers. 

\noindent\textbf{Transformer Decoder in LETR} is stacked with multiple decoder layers. Each decoder layer takes in a set of image features $\x' \in \mathbb{R}^{HW \times C}$ from the last encoder layer and a set of line entities $\mathbf{l} \in \mathbb{R}^{N \times C}$ from its predecessor decoder layer. The line entities are first processed with a SA module, each line entity $l \in \mathbb{R}^{C}$ in $\mathbf{l}$ attends to different regions of image feature embeddings $\x'$ via the CA module. FC layers and other modules are added into the pipeline similar to the Encoder setting above. %The whole pipeline can be described as SA(d)-Dropout-Skip-Norm-CA(d)-Dropout-Skip-Norm-FC(8d)-ReLU-Dropout-FC(d)-Skip-Norm. \\

\noindent{\em Line Entity Interpretation.} The \textit{line entities} are analogous with the \textit{object queries} in DETR \cite{carion2020end}. We found each line entity has its own preferred existing region, length, and orientation of potential line segment after the training process (shown in Figure \ref{fig:vis_q1}). We discuss line entities together make better predictions through self-attention and cross-attention refinement when encountering heterogeneous line segment structures in Section \ref{results_comparison} and Figure \ref{fig:qualitative}.

\subsection{Coarse-to-Fine Strategy}
\vspace{-2mm}
Different from object detection, line segment detection requires the detector to consider the local fine-grained details of line segments with the global indoor/outdoor structures together. In our LETR architecture, we propose a coarse-to-fine strategy to predict line segments in a refinement process. The process allows line entities to make precise predictions with the interaction of multi-scale encoded features while having an awareness of the holistic architecture with the communication to other line entities. During the coarse decoding stage, our line entities attend to potential line segment regions, often unevenly distributed, with a low resolution. During the fine decoding stage, our line entities produce detailed line segment predictions with a high resolution (see Figure \ref{fig:pipeline}). After each decoding layer at both coarse and fine decoding stage, we require line entities to make predictions through two shared prediction heads to make more precise predictions gradually.   

\noindent\textbf{Coarse Decoding.} 
During the coarse decoding stage, we pass image features and line entities into an encoder-decoder Transformer architecture. The encoder receives coarse features from the output of Conv5 (C5) from ResNet with $\frac{1}{32}$ original resolution. Then, line entity embeddings attend to coarse features from the output of the encoder in the cross-attention module at each layer. The coarse decoding stage is necessary for success at fine decoding stage and its high efficiency with less memory and computation cost. 

\noindent\textbf{Fine Decoding.}
The fine decoder inherits line entities from the coarse decoder and high-resolution features from the fine encoder. The features to the fine encoder come from the output of Conv4 (C4) from ResNet with $\frac{1}{16}$ original resolution. The line entity embeddings decode feature information in the same manner as the coarse decoding stage. 

\subsection{Line Segment Prediction}
% FIXME: 1) Add definition of line entities and their connection to queries
\vspace{-2mm}
In the previous decoding procedure, our multi-scale decoders progressively refine $N$ initial line entities to produce same amount final line entities. In the prediction stage. Each final entity $l$ will be fed into a feed-forward network (FFN), which consists of a classifier module to predict the confidence ${p}$ of being a line segment, and a regression module to predict the coordinates of two end points $\hat{\mathbf{p}}_1=(\hat{x}_1, \hat{y}_1)$, $\hat{\mathbf{p}}_2=(\hat{x}_2, \hat{y}_2)$ that parameterizes the associated line segment $\hat{\mathbf{L}}=(\hat{\mathbf{p}}_1, \hat{\mathbf{p}}_2)$. 
\vspace{-2mm}
\paragraph{Bipartite Matching.} Generally, there are many more line entities provided than actual line segments in the image. Thus, during the \textit{training} stage, we conduct a set-based bipartite matching between line segment predictions and ground-truth targets to determine whether the prediction is associated with an existing line segment or not: Assume there are $N$ line segment predictions $\{({p}^{(i)}, \hat{\mathbf{L}}^{(i)}); i=1,...,N\}$ and $M$ targets $\{\mathbf{L}^{(j)}; j=1,...,M\}$, we optimize a bipartite matching objective on a permutation function $\sigma(\cdot):\mathbb{Z}_+\rightarrow \mathbb{Z}_+$ which maps prediction indices $\{1,...,N\}$ to potential target indices $\{1,...,N\}$ (including $\{1,...,M\}$ for ground-truth targets and $\{M+1,...,N\}$ for unmatched predictions):
\vspace{-3mm}
\begin{align}
    % \displaystyle
    \mathcal{L}_{\text{match}} &= \sum_{i=1}^{N} \mathds{1}_{\{\sigma(i)\leq M\}} \big[ \lambda_1 d(\hat{\mathbf{L}}^{(i)}, {\mathbf{L}}^{(\sigma(i))}) - \lambda_2 {p}^{(i)}] \\
    \sigma^* &= \arg\min_{\sigma} \mathcal{L}_\text{match}
\end{align}
where $d(\cdot,\cdot)$ represents $L_1$ distance between coordinates and $\mathds{1}_{\{\cdot\}}$ is an indicator function. $\mathcal{L}_\text{match}$ takes both distance and confidence into account with balancing coefficients $\lambda_1,\lambda_2$. The optimal permutation $\sigma^*$ is computed using a Hungarian algorithm, mapping $M$ positive prediction indices to target indices $\{1,...,M\}$. During the \textit{inference} stage, we filter the $N$ line segment predictions by setting a fixed threshold on the confidence ${p}^{(i)}$ if needed due to no ground-truth provided.

\subsection{Line Segment Losses}
\vspace{-2mm}
We compute line segment losses based on the optimal permutation $\sigma^*$ from the bipartite matching procedure, in which $\{i;\sigma^*(i)\leq M\}$ represents indices of positive predictions.
\vspace{-2mm}
\paragraph{Classification Loss.} Based on a binary cross-entropy loss, we observe that hard examples are less optimized after learning rate decay and decide to apply adaptive coefficients inspired by focal loss \cite{FocalLoss17} to the classification loss term $\mathcal{L}_\text{cls}$:
\vspace{-3mm}
\begin{align}
\displaystyle
\mathcal{L}_\text{cls}^{(i)} =  
 & - \mathds{1}_{\{\sigma^*(i)\leq M\}} \alpha_1 (1- p^{(i)})^\gamma \log p^{(i)} \\ 
& - \mathds{1}_{\{\sigma^*(i)> M\}} \alpha_2 {p^{(i)}}^\gamma \log (1-p^{(i)}) 
\end{align}

\paragraph{Distance Loss.} We compute a simple $L_1$-based distance loss for line segment endpoint regression:
\begin{equation}
\displaystyle
\mathcal{L}_{\text{dist}}^{(i)} = \mathds{1}_{\{\sigma^*(i)\leq M\}}  d(\hat{\mathbf{L}}^{(i)}, {\mathbf{L}}^{(\sigma^*(i))}) 
\end{equation}
where $d(\cdot,\cdot)$ represents the sum of $L_1$ distances between prediction and target coordinates. The distance loss is only applied to the positive predictions. Note that we remove the GIoU loss from \cite{carion2020end} since GIoU is mainly designed for the similarity between bounding boxes instead of line segments. Thus, the final loss $\mathcal{L}$ of our model is formulated as:
\begin{equation}
    \mathcal{L} = \sum_{i=1}^N \lambda_\text{cls} \mathcal{L}_\text{cls}^{(i)} + \lambda_\text{dist} \mathcal{L}_\text{dist}^{(i)}
    \vspace{-3mm}
\end{equation}

\vspace{-2mm}
\section{Experiments}

\begin{figure*}
    \centering\hfill
    \begin{minipage}[t]{0.19\linewidth}\centering 
    \includegraphics[width=0.99\linewidth]{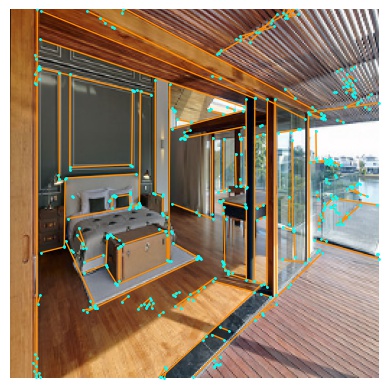} 
    \includegraphics[width=0.99\linewidth]{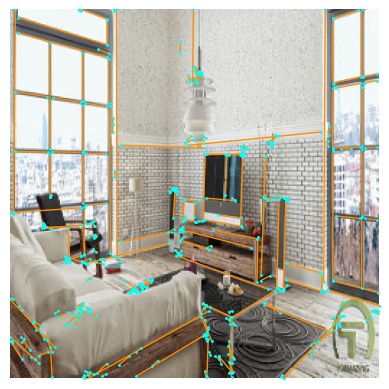} 
    \includegraphics[width=0.99\linewidth]{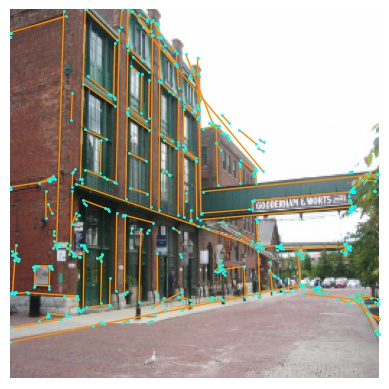} 
    \includegraphics[width=0.99\linewidth]{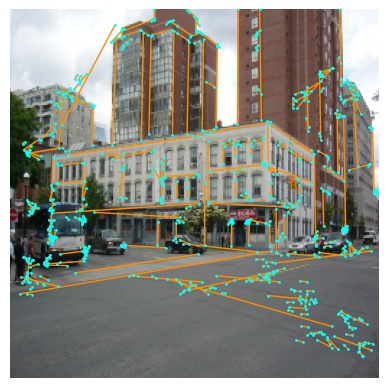} 
    {\small (a) AFM \cite{xue2019learning}}
    \end{minipage}\hfill
    \begin{minipage}[t]{0.19\linewidth}\centering
    \includegraphics[width=0.99\linewidth]{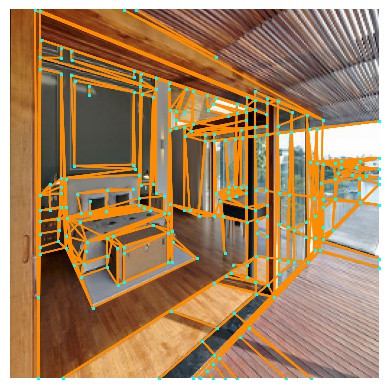} 
    \includegraphics[width=0.99\linewidth]{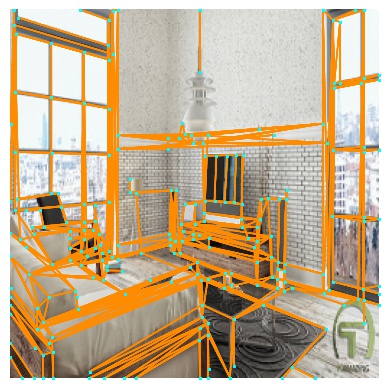} 
    \includegraphics[width=0.99\linewidth]{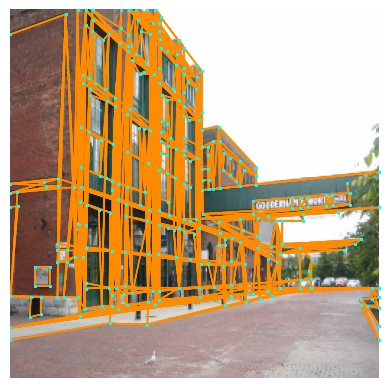}
    \includegraphics[width=0.99\linewidth]{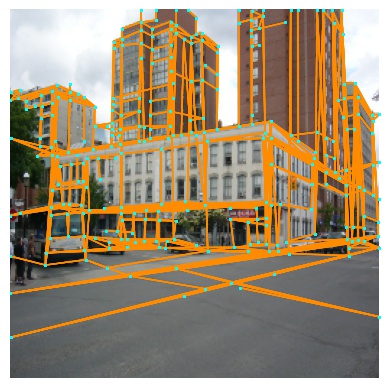}
    
    {\small (b) LCNN \cite{zhou2019end}}
    \end{minipage}\hfill
    \begin{minipage}[t]{0.19\linewidth}\centering
    \includegraphics[width=0.99\linewidth]{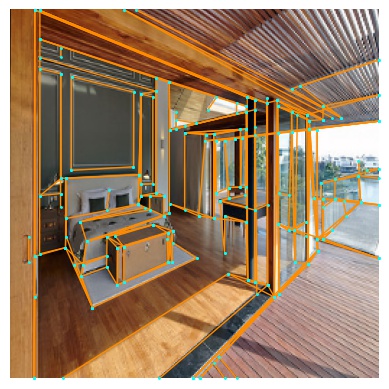}
    \includegraphics[width=0.99\linewidth]{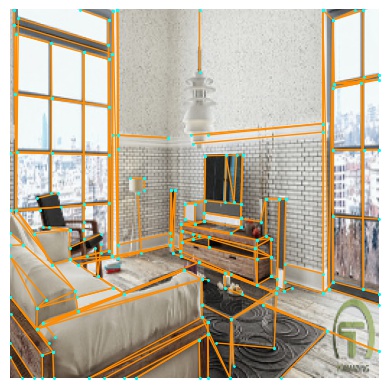}
    \includegraphics[width=0.99\linewidth]{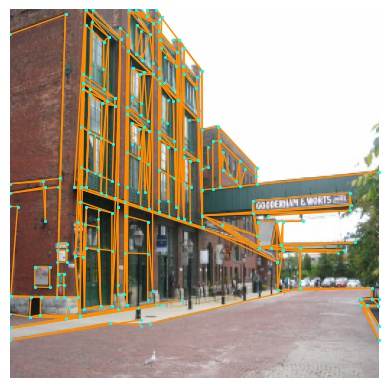}
    \includegraphics[width=0.99\linewidth]{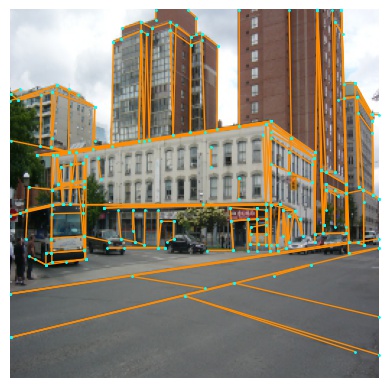}    
    
    {\small (c) HAWP \cite{xue2020holistically}}
    \end{minipage}\hfill
    \begin{minipage}[t]{0.19\linewidth}\centering
    \includegraphics[width=0.99\linewidth]{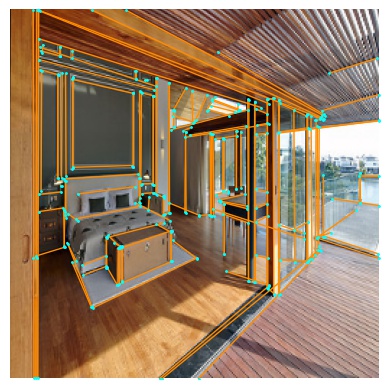}
    \includegraphics[width=0.99\linewidth]{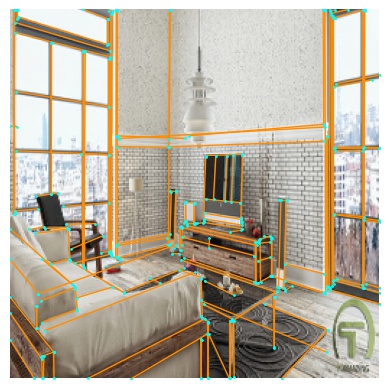}
    \includegraphics[width=0.99\linewidth]{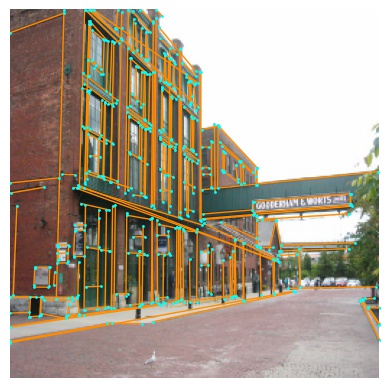}
    \includegraphics[width=0.99\linewidth]{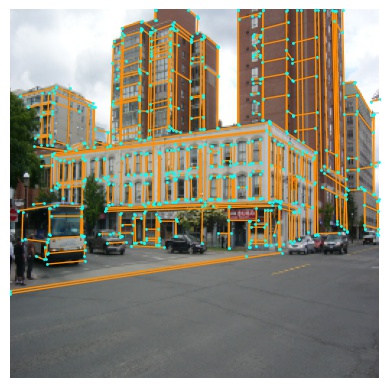}
    
    {\small (d) LETR (ours)}
    \end{minipage}\hfill
    \begin{minipage}[t]{0.19\linewidth}\centering
    \includegraphics[width=0.99\linewidth]{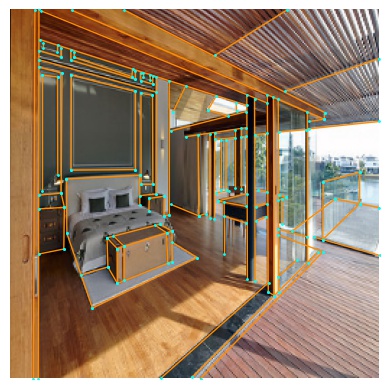}
    \includegraphics[width=0.99\linewidth]{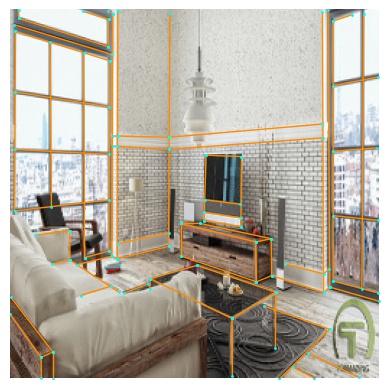}
    \includegraphics[width=0.99\linewidth]{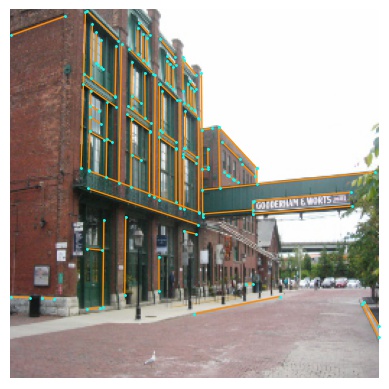}
    \includegraphics[width=0.99\linewidth]{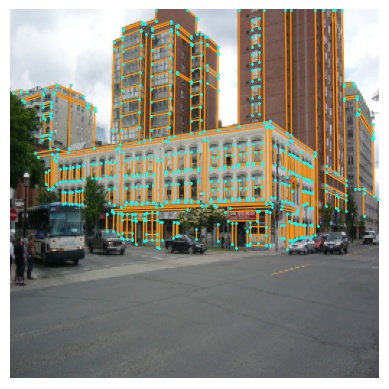}
    {\small (e) Ground-Truth}
    \end{minipage}
    \vspace{2mm}
    \caption{\textbf{Qualitative evaluation of line detection methods.} From left to right: the columns are the results from AFM \cite{xue2019learning},  LCNN \cite{zhou2019end}, HAWP \cite{xue2020holistically},  LETR (ours) and the ground-truth. From top to bottom: the top two rows are the results from the Wireframe test set, and the bottom two rows are the results from the YorkUrban test set.
    }
    \label{fig:qualitative}
\vspace{-2mm}
\end{figure*}

\subsection{Datasets}
We train and evaluate our model on the ShanghaiTech \textit{Wireframe} dataset \cite{huang2018learning}, which consists of 5000 training images and 462 testing images. We also evaluate our model on the \textit{YorkUrban} dataset \cite{denis2008efficient} with 102 testing images from both indoor scenes and outdoor scenes.

Through all experiments, we conduct data augmentations for the training set, including random horizontal/vertical flip, random resize, random crop, and image color jittering. At the training stage, we resize the image to ensure the shortest size is at least 480 and at most 800 pixels while the longest size is at most 1333. At the evaluation stage, we resize the image with the shortest side at least 1100 pixels.

\subsection{Implementation}
\paragraph{Networks.} We adopt both ResNet-50 and ResNet-101 as our feature backbone. For an input image $X \in \mathbb{R}^{ H_{0} \times W_{0} \times 3}$, the coarse encoder takes in the feature map from the Conv5 (C5) layer of ResNet backbone with resolution $x \in \mathbb{R}^{ H \times W \times C}$ where $H=\frac{H_{0}}{{32}}, W= \frac{W_{0}}{{32}}, C = 2048$. The fine encoder takes in a higher resolution feature map ($H=\frac{H_{0}}{{16}}, W=\frac{W_{0}}{{16}}, C = 1024$) from the Conv4 (C4) layer of ResNet. Feature maps are reduced to 256 channels by a 1x1 convolution and are fed into the Transformer along with the {\em sine/cosine} positional encoding. Our coarse-to-fine strategy consists of two independent encoder-decoder structures processing multi-scale image features. Each encoder-decoder structure is constructed with 6 encoder and 6 decoder layers with 256 channels and 8 attention heads.  
\vspace{-2mm}
\paragraph{Optimization.} We train our model using 4 Titan RTX GPUs through all our experiments. Model weights from DETR \cite{carion2020end} with ResNet-50 and ResNet-101 backbone are loaded as pre-training, and we discuss the effectiveness of pre-training in Section \ref{Pretraining}. We first train the coarse encoder-decoder for 500 epochs until optimal. Then, we freeze the weights in the coarse Transformer and train the fine Transformer initialized by coarse Transformer weights for 325 epochs (including a 25-epoch focal-loss fine-tuning). We adopt deep supervision \cite{lee2015deeply,xie2015holistically} for all decoder layers following DETR \cite{carion2020end}. FFN prediction head weights are shared through all decoder layers. We use AdamW as the model optimizer and set weight decay as $10^{-4}$. The initial learning rate is set to $10^{-4}$ and is reduced by a factor of 10 every 200 epochs for the coarse decoding stage and every 120 epochs for the fine prediction stage. We use 1000 line entities in all reported benchmarks unless specified elsewhere. To mitigate the class imbalance issue, we also reduce the classification weight for background/no-object instances by a factor of 10. 
\begin{table*}[!htp]
    \centering
    \vspace{1mm}
    \caption{\textbf{Comparison to prior work on Wireframe and YorkUrban benchmarks.} Our proposed LETR reaches state-of-the-art performance except sAP$^{10}$ and sAP$^{15}$ slightly worse than HAWP \cite{xue2020holistically} in Wireframe. FPS Results for LETRs are tested on a single Tesla V100. Results for other prior works are adopted from HAWP paper. }
      %\vspace{-3mm}
    \label{tab:benchmark}  
    \resizebox{0.95\linewidth}{!}{ 
    \begin{tabular}{|l|c|c|c|c|l|l|c|c|c|c|c|c|l|}
        \hline
        \multirow{2}{*}{Method} & \multicolumn{6}{c|}{\textit{Wireframe Dataset}} & \multicolumn{6}{c|}{\textit{YorkUrban Dataset}}  & FPS
        \\\cline{2-13}
        & sAP$^{10}$  & sAP$^{15}$  & sF$^{10}$ & sF$^{15}$  & AP$^{{H}}$ & F$^{{H}}$  & sAP$^{10}$  & sAP$^{15}$  & sF$^{10}$   & sF$^{15}$  & AP$^{{H}}$ & F$^{{H}}$ &
        \\\hline\hline
        LSD~\cite{VonGioi2010}  & /& / & /  & /& 55.2 & 62.5  & /& /& / & / & 50.9 & 60.1 & \textbf{49.6}\\\hline 
    
        DWP~\cite{huang2018learning}  &5.1& 5.9 & /& /  & 67.8  & 72.2  & 2.1 & 2.6 &/&/   & 51.0  & 61.6  & 2.24 \\\hline
    
        AFM~\cite{xue2019learning}   &24.4 & 27.5 & /& / & 69.2 & 77.2  & 9.4 &11.1 &/&/  & 48.2 & 63.3 & 13.5 \\\hline
        
        L-CNN~\cite{zhou2019end}  &62.9 & 64.9  & 61.3 & 62.4   & 82.8  & 81.3 & 26.4 & 27.5  & 36.9 &37.8    & 59.6 & 65.3 & 15.6   \\\hline
        HAWP~\cite{xue2020holistically}   &66.5 & 68.2  & 64.9 & 65.9  & 86.1  &83.1   & 28.5 & 29.7  & 39.7 & 40.5  & 61.2  & 66.3 & 29.5  \\\hline
        \textbf{LETR} (ours)  & 65.2 & 67.7  & \textbf{65.8} & \textbf{67.1}  & \textbf{86.3} & \textbf{83.3} & \textbf{29.4} & \textbf{31.7} & \textbf{40.1} & \textbf{41.8} & \textbf{62.7} & \textbf{66.9}  & 5.04 \\\hline
    \end{tabular}
    }
\end{table*}
\begin{figure*}[!htp]
\vspace{-3mm}
\begin{center}
\begin{tabular}{c}
\includegraphics[width=0.24\textwidth]{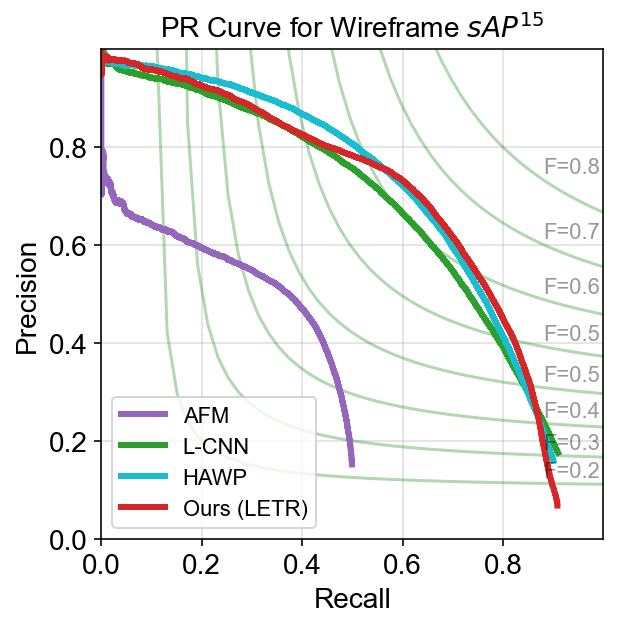}
\includegraphics[width=0.24\textwidth]{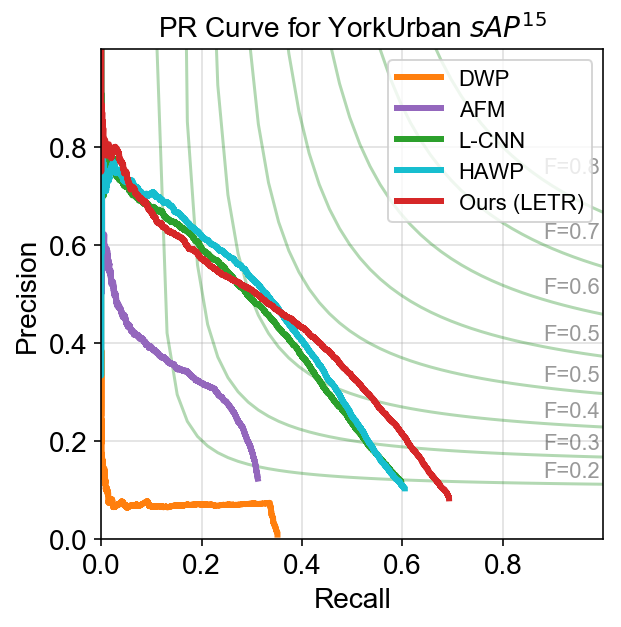}
\includegraphics[width=0.24\textwidth]{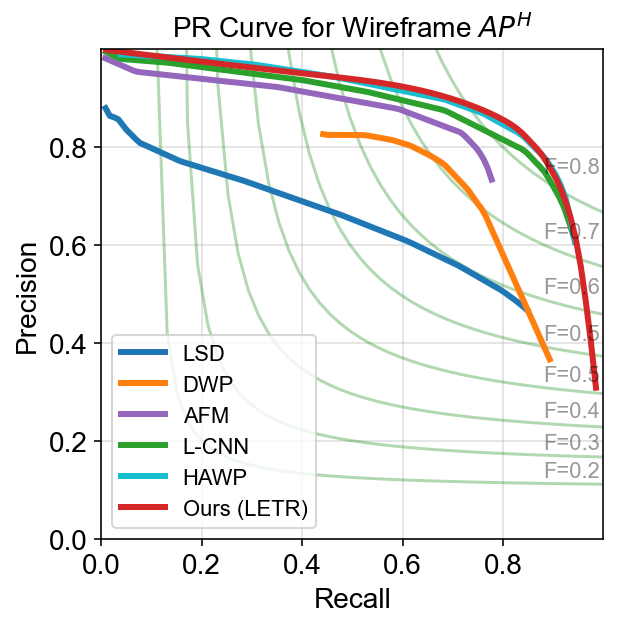}
\includegraphics[width=0.24\textwidth]{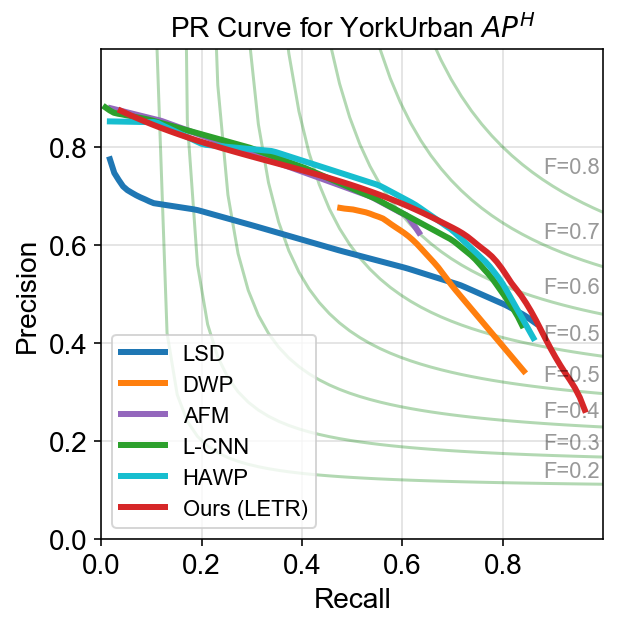}
\end{tabular}
\caption{\small \textbf{Precision-Recall (PR) curves.} PR curves of $\text{sAP}^{15}$ and $\text{AP}^{H}$ for DWP\cite{huang2018learning}, AFM\cite{xue2019learning}, L-CNN\cite{zhou2019end}, HAWP\cite{xue2020holistically} and LETR (ours) on Wireframe and YorkUrban benchmarks. }
\label{fig:PR}
\vspace{-8mm}
\end{center}
\end{figure*}

\subsection{Evaluation Metric}
\vspace{-2mm}
We evaluate our results based on two heatmap-based metrics, AP$^{H}$ and F$^{H}$, which are widely used in previous LSD task\cite{zhou2019end, huang2018learning}, and Structural Average Precision (sAP) which is proposed in L-CNN \cite{zhou2019end}. On top of that, we evaluate the result with a new metric, Structural F-score (sF), for a more comprehensive comparison. 

\noindent {\em Heatmap-based metrics, AP$^{H}$, F$^{H}$}: Prediction and ground truth lines are first converted to heatmaps by rasterizing the lines, and we generate the precision-recall curve comparing each pixel along with their confidence. Then we can use the curve to calculate F$^H$ and AP$^H$.

\noindent{\em Structural-based metrics, sAP\cite{zhou2019end}, sF:} Given a set of ground truth line and a set of predicted lines, for each ground-truth line $\mathbf{L}$, we define a predicted line $\hat{\mathbf{L}}$ to be a match of $\mathbf{L}$ if their $L_2$ distance is smaller than the pre-defined threshold $\vartheta \in \left\{10,15\right\}$. Over the set of lines matched to $\mathbf{L}$, we select the line with the highest confidence as a true positive and treat the rest as candidates for false positives. If the set of matching lines is empty, we would regard this ground-truth line as false negative. Each predicted line would be matched to at most one ground truth line, and if a line isn't matched to any ground-truth line, then it is considered as a false positive. The matching is recomputed at each confidence level to produce the precision-recall curve, and we consider sAP as the area under this curve. Considering $\text{F}^H$ as the complementary F-score measurement for $\text{AP}^H$, we evaluate the F-score measurement for sAP, denoted as sF, to be the best balanced performance measurement.

\subsection{Results and Comparisons}
\vspace{-2mm}
\label{results_comparison}

We summarize quantitative comparison results between LETR and previous line segment detection methods in Table \ref{tab:benchmark}. We report results for LETR with ResNet-101 backbone for Wireframe dataset and results with ResNet-50 backbone for York dataset. Our LETR achieves new state-of-the-art for all evaluation metrics on YorkUrban Dataset \cite{denis2008efficient}. In terms of heatmap-based evaluation metrics, our LETR is consistently better than other models for both benchmarks and outperforms HAWP \cite{xue2020holistically} by 1.5 for $\text{AP}^{H}$ on YorkUrban Dataset. We show PR curve comparison in Figure \ref{fig:PR} on $\text{sAP}^{15}$ and $\text{AP}^{H}$ for both Wireframe~\cite{huang2018learning} and YorkUrban benchmarks. In Figure \ref{fig:PR}, we notice the current limitation of LETR comes from lower precision prediction when we include fewer predictions compare to HAWP. When we include all sets of predictions, LETR predicts slightly better than HAWP and other leading methods, which matches our hypothesis that holistic prediction fashion can guide line entities to refine low confident predictions (usually due to local ambiguity and occlusion) with high confident predictions.

We also show both Wireframe and YorkUrban line segment detection qualitative results from LETR and other competing methods in Figure \ref{fig:qualitative}. The top two rows are indoor scene detection results from the Wireframe dataset, while the bottom two rows are outdoor scene detection results from the YorkUrban dataset.

\section{Ablation Study}
\vspace{-2mm}
\noindent\textbf{Compare with Object Detection Baselines.}
We compare LETR results with two object detection baseline where the line segments are treated as 2-D objects within this context in Table \ref{tab:compare-object-detection}. We see clear limitations for using bounding box diagonal for both Faster R-CNN and DETR responding to our motivation in Section \ref{motivation}. 

\begin{table}[h]
    \centering
    \vspace{-2mm}
    \caption{\textbf{Comparison with object detection baselines} on Wireframe~\cite{huang2018learning}.}
    \label{tab:compare-object-detection}

    \resizebox{0.7\linewidth}{!}{ 
    \begin{tabular}{|l|c|c|c|c|}
        \hline
        {Method} &  sAP$^{10}$ &  sAP$^{15}$   & sF$^{10}$   & sF$^{15}$ 
        \\\hline
        Faster R-CNN  & 38.4 & 40.7 & 51.5 & 53.0 \\\hline
        Vanilla DETR  & 53.8 & 57.2  & 57.2 & 59.0 \\\hline
        LETR (ours)  & 65.2 & 67.7  & 65.8 & 67.1 \\\hline
    \end{tabular}
    }
    \vspace{-1mm}
\end{table}

\noindent\textbf{Effectiveness of Multi-Stage Training.}
We compare the effectiveness of different modules in LETR in Table \ref{table:role-of-modules}. During the coarse decoding stage, LETR reaches 62.3 and 65.2 for sAP$^{10}$ and sAP$^{15}$ with encoding features from the C5 layer of ResNet backbone, and 63.8 and 66.5 with the one from C4 of ResNet backbone. The fine decoder reaches 64.7 and 67.4 for sAP$^{10}$ and sAP$^{15}$ by improving the coarse prediction with fine-grained details from high-resolution features. We then adjust the data imbalance problem with focal loss to reach 65.2 and 67.7 for sAP$^{10}$ and sAP$^{15}$. 

\begin{figure}[!ht]
\begin{center}
\vspace{-2mm}
\begin{tabular}{c}
\hspace{-4mm}\includegraphics[width=0.16\textwidth]{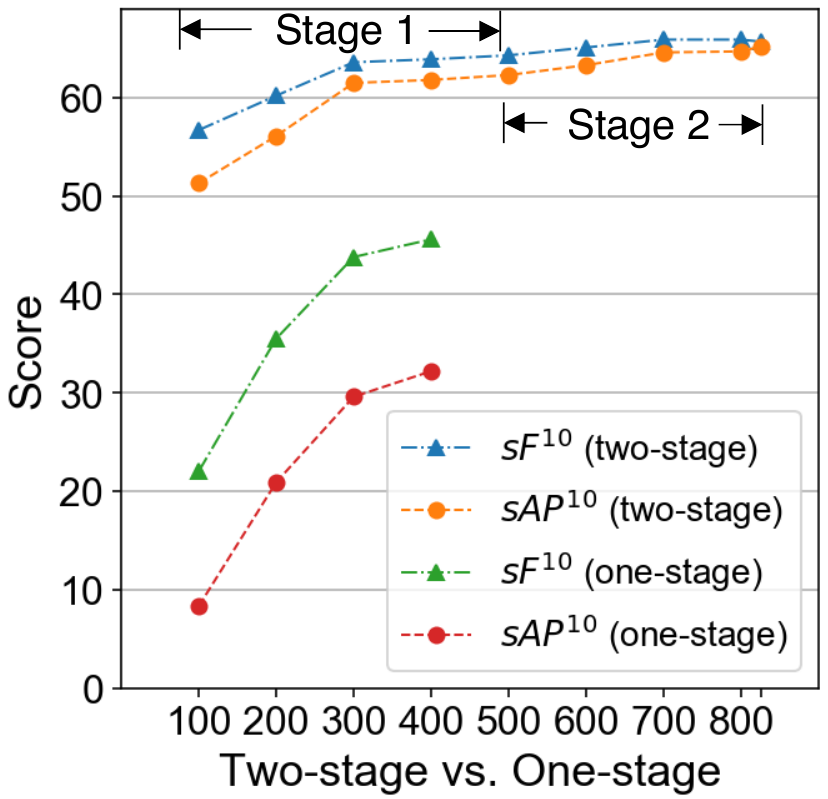}
\includegraphics[width=0.16\textwidth]{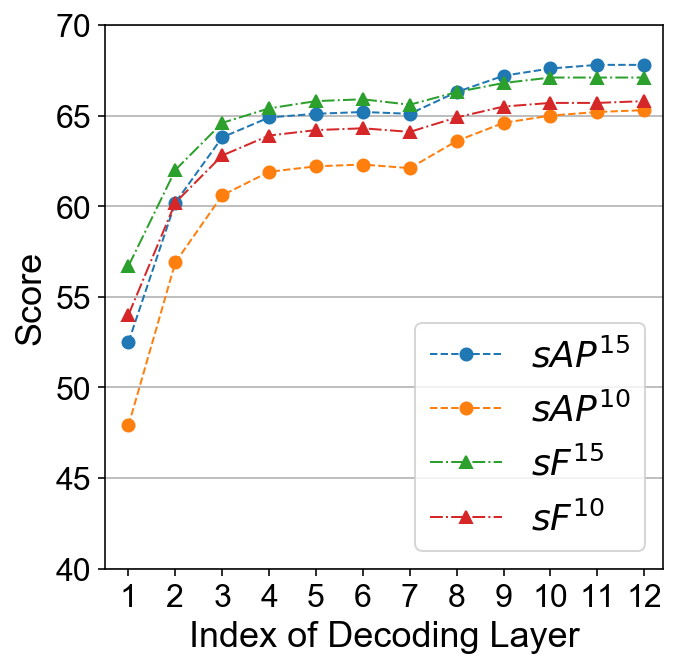}
\includegraphics[width=0.16\textwidth]{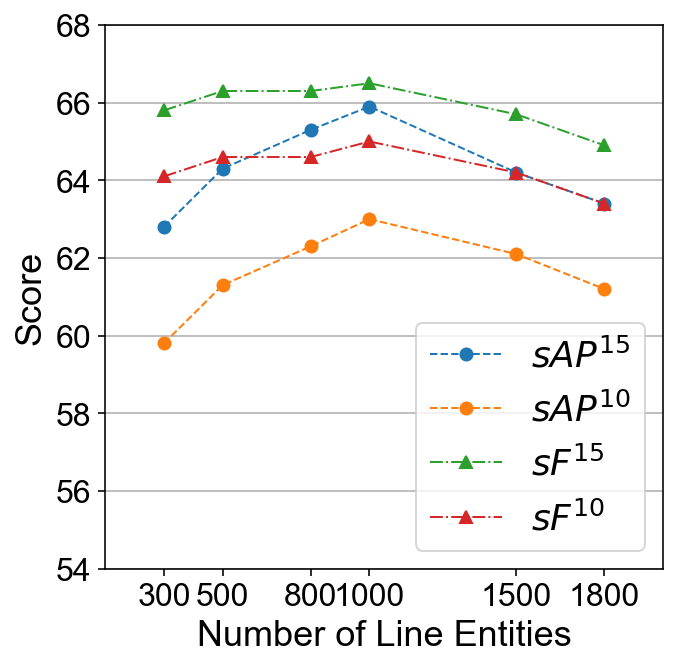}
\end{tabular}
\caption{\small (a) \textbf{Multi-stage vs. single-stage training.} We compare results training coarse and fine layers in single stages and multi-stages (b) \textbf{Number of decoding layers.} We evaluate the performance of outputs from each decoding layer. The 1-6 layers are coarse decoder layers and 7-12 layers fine decoder layers. (c) \textbf{Number of line entities.} We test LETR (coarse decoding stage only) with different numbers of line entities on Wireframe.}
\vspace{-6mm}
\label{fig:ablation}
\end{center}
\end{figure}

As shown in Figure \ref{fig:ablation}~(a), we found it is necessary to train the fine decoding stage after the coarse decoding stage converges. Training both stages together as a one-stage model results a significant worse performance after 400 epochs. 

\noindent\textbf{Effect of Number of Queries.}
We found a large number of line entities is essential to the line segment detection task by experimenting on a wide range of the number of line entities (See Figure \ref{fig:ablation}~(c), and using 1000 line entities is optimal for the Wireframe benchmark which contains 74 line segments in average.

\begin{table}[!ht]
\centering
\vspace{-2mm}
\caption{ \small{\textbf{Effectiveness of modules.}} Ablation study of the architecture design and learning aspects in the proposed LETR on Wireframe dataset. (C) indicates the indexed feature used for coarse decoder; (F) indicates the indexed feature used for fine decoder.}
\label{table:role-of-modules}
\scalebox{0.7}{
\begin{small}
\begin{tabular}{|cccc|cc|}
    \hline
    \textbf{Coarse Decoding} & \textbf{Fine Decoding} & \textbf{Focal Loss} & \textbf{Feature Index}
     &  \textbf{sAP$^{10}$ } & \textbf{sAP$^{15}$ } 
    \\ [0.5ex] 
    \hline 
    \checkmark &  &    &  C5(C) & 62.3 &  65.2 \\ 
    \checkmark &  &    &  C4(C)  & 63.8 & 66.5 \\
    \checkmark & \checkmark &   & C5(C), C4(F) & 64.7 & 67.4 \\
    \checkmark & \checkmark & \checkmark & C5(C), C4(F) & 65.2  &  67.7 \\   
    \hline
\end{tabular}
\end{small}
}
\vspace{-3mm}
\end{table}

\noindent\textbf{Effect of Image Upsampling.}
All algorithms see the same input image resolution (640$\times$480 typically). However, some algorithms try more precise predictions by upsampling images. To understand the impact of upsampling, we train and test HAWP and LETR under multiple upsampling scales. In Table \ref{tab:image-size} below, higher training upsampling resolution improves both methods. LETR obtains additional gains with higher test upsampling resolution. 

\begin{table}[h]
    \centering
    \vspace{-2mm}
    \caption{\textbf{Effectiveness of upsampling} with Wireframe dataset. LETR uses ResNet-101 backbone. * Our LETR-512 resizes original image with the shortest size in a range between 288 and 512 $\dagger$ Our LETR-800 resizes original image with the shortest size in a range between 480 and 800.}
    \label{tab:image-size}
    %\vspace{-3mm}
    \resizebox{0.9\linewidth}{!}{ 
    \begin{tabular}{|c|c|c|c|c|c|c|}
        \hline
        & Train Size & Test Size & $\text{sAP}^{10}$ & $\text{sAP}^{15}$ & $\text{sF}^{10}$ & $\text{sF}^{15}$  
        \\\hline\hline
        HAWP & 512  & 512  & 65.7  & 67.4 & 64.7 & 65.8 \\\hline
        HAWP & 832  & 832  & \textbf{67.7}  & \textbf{69.1} & 65.5 & 66.4 \\\hline
        HAWP & 832  & 1088 & 65.7 & 67.1 & 64.3 & 65.1 \\\hline
        LETR & 512*  & 512  & 61.1 & 64.1 & 63.1 & 64.8 \\\hline
        LETR & 800$\dagger$  & 800  & 64.3  & 67.0 & 65.5 & 66.9 \\\hline
        LETR & 800$\dagger$  & 1100 & 65.2  & 67.7 & \textbf{65.8} & \textbf{67.1}  \\\hline
    \end{tabular}
    }
    %\end{minipage}
    \vspace{-1mm}
\end{table}

\noindent\textbf{Effectiveness of Pretraining.}\label{Pretraining}
We found model pretraining is essential for LETR to obtain state-of-the-art results. With DETR pretrained weights for COCO object detection \cite{lin2014microsoft}, our coarse-stage-only model converges at 500 epochs. With CNN backbone pretrained weights for ImageNet classification, our coarse-stage-only model converges to a lower score at 900 epochs. Without pretraining, LETR is difficult to train due to the limited amount of data in the Wireframe benchmark.   

\begin{table}[h]
    \centering
    \vspace{-2mm}
    \caption{\textbf{Effectiveness of pretraining.} We train LETR (coarse decoding stage only) with two variants. ImageNet represents LETR with ImageNet pretrained ResNet backbone. COCO represents LETR with COCO pretrained DETR weights.}
    \vspace{1mm}
    \resizebox{0.8\linewidth}{!}{ 
    \begin{tabular}{|l|c|c|c|c|c|}
        \hline
        {Method} & Epochs &  sAP$^{10}$ &  sAP$^{15}$   & sF$^{10}$   & sF$^{15}$ 
        \\\hline
        ImageNet & 900 & 58.4 & 62.0 & 62.4 & 64.6 \\\hline
        COCO     & 500 & 62.3 & 65.2 & 64.3 & 65.9 \\\hline
    \end{tabular}
    }
    \vspace{-5mm}
    \label{tab:compare-pretraining}
\end{table}

\vspace{-2mm}
\section{Visualization}
\label{vis_section}
\begin{figure}[!htp]
\centering
\vspace{-3mm}
\begin{tabular}{c}
\includegraphics[width=0.45\textwidth]{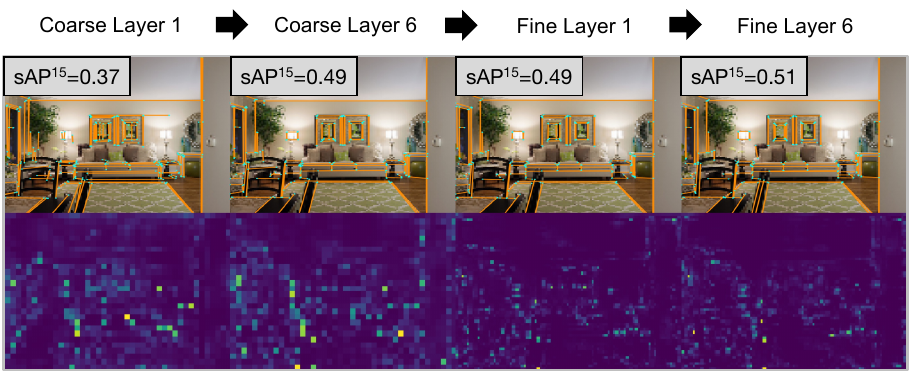} \\
\end{tabular}
\caption{\small \textbf{Visualization of LETR coarse-to-fine decoding process.} From top to bottom: The $1^{st}$  row shows line segment detection results based on line entities after different layers and the $2^{nd}$ row shows its corresponding overlay of attention heatmaps. From left to right: The $1^{st}$, $2^{nd}$, $3^{rd}$, $4^{th}$ columns are coarse decoder layer 1, coarse decoder layer 6, fine decoder layer 1, fine decoder layer 6, respectively.} 
\label{fig:vis_coarse_to_fine}
\vspace{-3mm}
\end{figure}

\vspace{-2mm}
We demonstrate LETR's coarse-to-fine decoding process in Figure \ref{fig:vis_coarse_to_fine}. The first two columns are results from the coarse decoder receiving decoded features from the C5 ResNet layer. While the global structure of the scene is well-captured efficiently, the low-resolution features prevent it from making predictions precisely. The last two columns are results from the fine decoder receiving decoded features from the C4 ResNet layer and line entities from the coarse decoder. The overlay of attention heatmaps depicts more detailed relations in the image space, which is the key to the detector performance. This finding is also shown in Figure \ref{fig:ablation}(b), where the decoded output after each layer has consistent improvement with the multi-scale encoder-decoder strategy.

\section{Conclusion}
\vspace{-2mm}
In this paper, we presented LETR, a line segment detector based on a multi-scale encoder/decoder Transformer structure. By casting the line segment detection problem in a holistically end-to-end fashion, we perform set prediction without explicit edge/junction/region detection and heuristics-guided perceptual grouping processes. A direct endpoint distance loss allows geometric structures beyond bounding box representations to be modeled and predicted.

\vspace{1mm}
\noindent{\bf Acknowledgment}. \small{This work is funded by NSF IIS-1618477 and NSF IIS-1717431. We thank Justin Lazarow, Feng Han, Ido Durst, Yuezhou Sun, Haoming Zhang, and Heidi Cheng for valuable feedbacks.}

{\small
\bibliographystyle{ieee_fullname}
\bibliography{egbib}

\begin{thebibliography}{10}\itemsep=-1pt

\bibitem{boldt1989token}
Michael Boldt, Richard Weiss, and Edward Riseman.
\newblock Token-based extraction of straight lines.
\newblock {\em IEEE Transactions on Systems, Man, and Cybernetics},
  19(6):1581--1594, 1989.

\bibitem{BurnsHR86}
J.~Brian Burns, Allen~R. Hanson, and Edward~M. Riseman.
\newblock Extracting straight lines.
\newblock {\em IEEE Trans. Pattern Anal. Mach. Intell.}, 8(4):425--455, 1986.

\bibitem{canny1986computational}
John Canny.
\newblock A computational approach to edge detection.
\newblock {\em IEEE Transactions on pattern analysis and machine intelligence},
  (6):679--698, 1986.

\bibitem{carion2020end}
Nicolas Carion, Francisco Massa, Gabriel Synnaeve, Nicolas Usunier, Alexander
  Kirillov, and Sergey Zagoruyko.
\newblock End-to-end object detection with transformers.
\newblock In {\em Eur. Conf. Comput. Vis.}, 2020.

\bibitem{denis2008efficient}
Patrick Denis, James~H Elder, and Francisco~J Estrada.
\newblock Efficient edge-based methods for estimating manhattan frames in urban
  imagery.
\newblock In {\em European conference on computer vision}, 2008.

\bibitem{devlin2018bert}
Jacob Devlin, Ming-Wei Chang, Kenton Lee, and Kristina Toutanova.
\newblock Bert: Pre-training of deep bidirectional transformers for language
  understanding.
\newblock In {\em NAACL-HLT}, 2019.

\bibitem{dollar2006supervised}
Piotr Doll{\'a}r, Zhuowen Tu, and Serge Belongie.
\newblock Supervised learning of edges and object boundaries.
\newblock In {\em IEEE Conf. Comput. Vis. Pattern Recog.}, 2006.

\bibitem{dollar2013structured}
Piotr Doll{\'a}r and C~Lawrence Zitnick.
\newblock Structured forests for fast edge detection.
\newblock In {\em IEEE Conf. Comput. Vis. Pattern Recog.}, 2013.

\bibitem{duda1972use}
Richard~O Duda and Peter~E Hart.
\newblock Use of the hough transformation to detect lines and curves in
  pictures.
\newblock {\em Communications of the ACM}, 15(1):11--15, 1972.

\bibitem{elder2002ecological}
James~H Elder and Richard~M Goldberg.
\newblock Ecological statistics of gestalt laws for the perceptual organization
  of contours.
\newblock {\em Journal of Vision}, 2(4):5--5, 2002.

\bibitem{fu2019dual}
Jun Fu, Jing Liu, Haijie Tian, Yong Li, Yongjun Bao, Zhiwei Fang, and Hanqing
  Lu.
\newblock Dual attention network for scene segmentation.
\newblock In {\em Proceedings of the IEEE Conference on Computer Vision and
  Pattern Recognition}, pages 3146--3154, 2019.

\bibitem{furukawa2003accurate}
Yasutaka Furukawa and Yoshihisa Shinagawa.
\newblock Accurate and robust line segment extraction by analyzing distribution
  around peaks in hough space.
\newblock {\em Computer Vision and Image Understanding}, 92(1):1--25, 2003.

\bibitem{guil1995fast}
Nicolas Guil, Julio Villalba, and Emilio~L Zapata.
\newblock A fast hough transform for segment detection.
\newblock {\em IEEE Transactions on Image Processing}, 4(11):1541--1548, 1995.

\bibitem{he2016deep}
Kaiming He, Xiangyu Zhang, Shaoqing Ren, and Jian Sun.
\newblock Deep residual learning for image recognition.
\newblock In {\em CVPR}, 2016.

\bibitem{huang2018learning}
Kun Huang, Yifan Wang, Zihan Zhou, Tianjiao Ding, Shenghua Gao, and Yi Ma.
\newblock Learning to parse wireframes in images of man-made environments.
\newblock In {\em IEEE Conf. Comput. Vis. Pattern Recog.}, pages 626--635,
  2018.

\bibitem{krizhevsky2012imagenet}
Alex Krizhevsky, Ilya Sutskever, and Geoffrey~E Hinton.
\newblock {ImageNet} classification with deep convolutional neural networks.
\newblock {\em Adv. Neural Inform. Process. Syst.}, 2012.

\bibitem{lee2015deeply}
Chen-Yu Lee, Saining Xie, Patrick Gallagher, Zhengyou Zhang, and Zhuowen Tu.
\newblock Deeply-supervised nets.
\newblock In {\em Artificial intelligence and statistics}, pages 562--570,
  2015.

\bibitem{FocalLoss17}
Tsung{-}Yi Lin, Priya Goyal, Ross~B. Girshick, Kaiming He, and Piotr
  Doll{\'{a}}r.
\newblock Focal loss for dense object detection.
\newblock In {\em Int. Conf. Comput. Vis.}, pages 2999--3007, 2017.

\bibitem{lin2014microsoft}
Tsung-Yi Lin, Michael Maire, Serge Belongie, James Hays, Pietro Perona, Deva
  Ramanan, Piotr Doll{\'a}r, and C~Lawrence Zitnick.
\newblock Microsoft coco: Common objects in context.
\newblock In {\em Eur. Conf. Comput. Vis.}, pages 740--755. Springer, 2014.

\bibitem{long2015fully}
Jonathan Long, Evan Shelhamer, and Trevor Darrell.
\newblock Fully convolutional networks for semantic segmentation.
\newblock {\em IEEE Conf. Comput. Vis. Pattern Recog.}, 2015.

\bibitem{lu2015cannylines}
Xiaohu Lu, Jian Yao, Kai Li, and Li Li.
\newblock Cannylines: A parameter-free line segment detector.
\newblock In {\em 2015 IEEE International Conference on Image Processing
  (ICIP)}, pages 507--511. IEEE, 2015.

\bibitem{marr1982vision}
David Marr.
\newblock Vision: A computational investigation into the human representation
  and processing of visual information, henry holt and co.
\newblock {\em Inc., New York, NY}, 2(4.2), 1982.

\bibitem{martin2004learning}
David~R Martin, Charless~C Fowlkes, and Jitendra Malik.
\newblock Learning to detect natural image boundaries using local brightness,
  color, and texture cues.
\newblock {\em IEEE transactions on pattern analysis and machine intelligence},
  26(5):530--549, 2004.

\bibitem{matas2000robust}
Jiri Matas, Charles Galambos, and Josef Kittler.
\newblock Robust detection of lines using the progressive probabilistic hough
  transform.
\newblock {\em Computer vision and image understanding}, 78(1):119--137, 2000.

\bibitem{nieto2011line}
Marcos Nieto, Carlos Cuevas, Luis Salgado, and Narciso Garc{\'\i}a.
\newblock Line segment detection using weighted mean shift procedures on a 2d
  slice sampling strategy.
\newblock {\em Pattern Analysis and Applications}, 14(2):149--163, 2011.

\bibitem{fasterrcnn}
Shaoqing Ren, Kaiming He, Ross Girshick, and Jian Sun.
\newblock Faster r-cnn: Towards real-time object detection with region proposal
  networks.
\newblock In {\em Advances in neural information processing systems}, pages
  91--99, 2015.

\bibitem{smith1997susan}
Stephen~M Smith and J~Michael Brady.
\newblock Susan—a new approach to low level image processing.
\newblock {\em International journal of computer vision}, 23(1):45--78, 1997.

\bibitem{tu2008auto}
Zhuowen Tu.
\newblock Auto-context and its application to high-level vision tasks.
\newblock In {\em IEEE Conference on Computer Vision and Pattern Recognition},
  2008.

\bibitem{vaswani2017attention}
Ashish Vaswani, Noam Shazeer, Niki Parmar, Jakob Uszkoreit, Llion Jones,
  Aidan~N Gomez, {\L}ukasz Kaiser, and Illia Polosukhin.
\newblock Attention is all you need.
\newblock In {\em Advances in neural information processing systems}, pages
  5998--6008, 2017.

\bibitem{VonGioi2010}
R~G von Gioi, J Jakubowicz, J~M Morel, and G Randall.
\newblock {LSD: A Fast Line Segment Detector with a False Detection Control}.
\newblock {\em IEEE Trans. Pattern Anal. Mach. Intell.}, 32(4):722--732, 2010.

\bibitem{wang2018non}
Xiaolong Wang, Ross Girshick, Abhinav Gupta, and Kaiming He.
\newblock Non-local neural networks.
\newblock In {\em Proceedings of the IEEE conference on computer vision and
  pattern recognition}, pages 7794--7803, 2018.

\bibitem{xie2015holistically}
Saining Xie and Zhuowen Tu.
\newblock Holistically-nested edge detection.
\newblock In {\em Proceedings of the IEEE international conference on computer
  vision}, pages 1395--1403, 2015.

\bibitem{xue2019learning}
Nan Xue, Song Bai, Fudong Wang, Gui-Song Xia, Tianfu Wu, and Liangpei Zhang.
\newblock Learning attraction field representation for robust line segment
  detection.
\newblock In {\em IEEE Conf. Comput. Vis. Pattern Recog.}, 2019.

\bibitem{xue2020holistically}
Nan Xue, Tianfu Wu, Song Bai, Fudong Wang, Gui-Song Xia, Liangpei Zhang, and
  Philip~HS Torr.
\newblock Holistically-attracted wireframe parsing.
\newblock In {\em Proceedings of the IEEE/CVF Conference on Computer Vision and
  Pattern Recognition}, pages 2788--2797, 2020.

\bibitem{zhang2019self}
Han Zhang, Ian Goodfellow, Dimitris Metaxas, and Augustus Odena.
\newblock Self-attention generative adversarial networks.
\newblock In {\em International Conference on Machine Learning}, pages
  7354--7363. PMLR, 2019.

\bibitem{PPGNet}
Ziheng Zhang, Zhengxin Li, Ning Bi, Jia Zheng, Jinlei Wang, Kun Huang, Weixin
  Luo, Yanyu Xu, and Shenghua Gao.
\newblock Ppgnet: Learning point-pair graph for line segment detection.
\newblock In {\em IEEE Conf. Comput. Vis. Pattern Recog.}, 2019.

\bibitem{zhou2019end}
Yichao Zhou, Haozhi Qi, and Yi Ma.
\newblock End-to-end wireframe parsing.
\newblock In {\em Proceedings of the IEEE International Conference on Computer
  Vision}, pages 962--971, 2019.

\end{thebibliography}
}

\end{document}